\documentclass[journal]{IEEEtran}
\usepackage{blindtext}
\usepackage{graphicx}
\usepackage{amsmath}
\usepackage{amsfonts}
\usepackage{amsthm}
\usepackage{amssymb}
\usepackage{multirow}
\usepackage{gensymb}
\usepackage[pdftex]{hyperref}
\usepackage{caption}
\usepackage{bm}
\usepackage{algorithm,algpseudocode}

\hypersetup{colorlinks=true}

\ifCLASSINFOpdf
\else
\fi

\begin{document}

\title{A Recursive Least Square Method for 3D Pose Graph Optimization Problem}

\author{S.M.~Nasiri,~\IEEEmembership{Student Member}, IEEE, 
        Reshad~Hosseini,
        and~Hadi~Moradi,~\IEEEmembership{Senior Member}, IEEE
}


\maketitle

\begin{abstract}
Pose Graph Optimization (PGO) is an important non-convex optimization problem and is the state-of-the-art formulation for SLAM in robotics. It also has applications like camera motion estimation, structure from motion and 3D reconstruction in machine vision.
Recent researches have shown the importance of good initialization to bootstrap well-known iterative PGO solvers to converge to good solutions. The state-of-the-art initialization methods, however, works in low noise or eventually moderate noise problems, and they fail in challenging problems with high measurement noise. Consequently, iterative methods may get entangled in local minima in high noise scenarios.

In this paper we present an initialization method\footnote{A preliminary version of this work appeared at the International Conference on Robotics and Automation (ICRA 2018), wherein our initialization algorithm was originally introduced.} which uses orientation measurements and then present a convergence analysis of our iterative algorithm. We show how the algorithm converges to global optima in noise-free cases and also obtain a bound for the difference between our result and the optimum solution in scenarios with noisy measurements. We then present our second algorithm that uses both relative orientation and position measurements to obtain a more accurate least squares approximation of the problem that is again solved iteratively.

In the convergence proof, a structural coefficient arises that has important influence on the basin of convergence. Interestingly, simulation results show that this coefficient also affects the performance of other solvers and so it can indicate the complexity of the problem. Experimental results show the excellent performance of the proposed initialization algorithm, specially in high noise scenarios.
%
\end{abstract}


\IEEEpeerreviewmaketitle
\newtheorem{Proposition}{Proposition}
\newcommand{\argmin}{\mathop{\mathrm{argmin}}}
\newcommand{\argmax}{\mathop{\mathrm{argmax}}}
\newcommand{\vs}{\mathbf{s}}
\newcommand{\vm}{\mathbf{m}}
\newcommand{\vc}{\mathbf{c}}
\newcommand{\vb}{\mathbf{b}}
\newcommand{\va}{\mathbf{a}}
\newcommand{\vd}{\mathbf{d}}
\newcommand{\ve}{\mathbf{e}}
\newcommand{\vf}{\mathbf{f}}
\newcommand{\vz}{\mathbf{z}}
\newcommand{\vl}{\mathbf{l}}
\newcommand{\vp}{\mathbf{p}}
\newcommand{\del}{\bm{\delta}}

\section{Introduction}\label{Sec:Intro}
Pose Graph Optimization (PGO) is the problem of finding robot poses such that relative poses would be the best description of measurements. The PGO is the most applied formulation of Simultaneous Localization And Mapping (SLAM) in robotics. It is also known as SE($d$)-Synchronization in many other fields.

The PGO problem is a non-convex optimization problem and hard to solve in general. The iterative gradient-based methods employed for solving the PGO problem cannot guarantee the global optimality of the solution. Well-known iterative numerical methods were used to solve PGO problem in the literature, such as Guess-Newton \cite{lu1997globally}, \cite{kummerle2011g}, \cite{kaess2008isam}, Levenberg-Marquardt \cite{olson2006fast} and trust region \cite{rosen2012incremental}, \cite{rosen2014rise}. Convergence behavior of Guess-Newton method in PGO problem was investigated by L. Carlone in \cite{carlone2013convergence}. Grisetti et al. \cite{grisetti2009nonlinear} presented method that have larger basins of convergence. But L. Carlone et al. \cite{carlone2014fast}, \cite{carlone2014angular} and D. Rosen et al. \cite{rosen2015convex} showed that state-of-the-art algorithms are entangled in local minima and failed to converge to the global optimum. Lagrangian duality formulation is developed and used for 2D SLAM in \cite{carlone2015duality} and was extended to 3D SLAM in \cite{carlone2015lagrangian}.

Iterative methods start from an initial guess and refine the solution in each step. Depending on the graph structure and the noise level of measurements, the problem can have different local minima that are far from the global minimum. Therefore choosing a good initial point can lead to faster convergence, and reduce the risk of convergence to a local minimum. Many Initialization methods \cite{carlone2014fast}, \cite{carlone2014angular}, \cite{carlone2012linear}, \cite{martinec2007robust} used this fact that a part of the PGO cost function that is related to orientation measurements is independent from positions and can be solved separately. Then the orientations can be assumed to be constant to calculate positions from a remaining least squares cost function. This solution can be used as a good initialization to bootstrap iterative methods.

L. Carlone et al. \cite{carlone2015initialization} reviewed and compared several initialization methods \cite{peters2015sensor}, \cite{martinec2007robust}, \cite{hartley2013rotation}, \cite{fredriksson2012simultaneous}, \cite{tron2014distributed} and showed that the chordal initialization has the best performance in common benchmark scenarios. In a recent work \cite{briales2017initialization}, the authors used Lagrangian relaxation to give a good initialization. Their techniques uses both rotational and translational information.

In this paper, we will present a method that guesses the orientations of the PGO problem and then positions are computed as a result of a least-squares problem. Subsequently, a convergence analysis will be conducted in order to show that our method converges to optimal solution iteratively in noise free cases. Also a convergence analysis will be performed for noisy cases, which obtains a boundary for the distance between the output of our algorithm and the optimal solution. Simulation results show that the proposed method give a better initialization than chordal methods and can help iterative solvers such as well-known Gauss-Newton and Levenberg-Marquardt to reach a more reliable solution. Also we will propose another algorithm which uses both orientations and positions and approximates the PGO cost function by a quadratic cost function that solve iteratively using least-squares solver. Experimental results confirm that for low and moderate measurements noise a few iterations of our algorithm will give a solution that is very close to the optimal solution. Since proposed methods iteratively approximates the problem by a linear least-squares problem, they are computationally efficient and very fast. We will show that using our algorithms as the initializer in scenarios with high level of noise, can improve the results of the state-of-the-art SE-Sync method \cite{rosen2017se}. We also introduce a structural coefficient that affects the basin of convergence of our algorithms. Simulation results also confirm the effect of this coefficient on the performance of other solvers.


In the following, we will first explain the basic mathematic notations and preliminaries in \S\ref{Sec:Note}. We will then propose our method that guesses the orientations of the PGO problem in \S\ref{Sec:LS} and \S\ref{Sec:Alg}. The convergence analysis will be presented for scenarios with noise-free (\S\ref{Sec:CA}-A) and noisy measurements(\S\ref{Sec:CA}-B). The second algorithm which uses both orientations and positions will be presented in \S\ref{Sec:PGOLS}. And finally, several experimental results for common scenarios with various amount of noise presented in \S\ref{Sec:Eval} to show the performance of our algorithms.


\section{Notations and Preliminaries}\label{Sec:Note}
In this part the notations and basic mathematics are explained.
\subsection{PGO Problem and Graph Topology}
A set of poses in 3-dimensional space is denoted by $n+1$ positions $\mathbf{p}_0,...,\mathbf{p}_n$ and $n+1$ orientations $R_0,...,R_n$. The relative noisy displacement and orientation measurements from pose $i$ to pose $j$ are denoted by $\mathbf{d}_{ij}$ and $Z_i^j$, respectively. PGO is the problem of estimating poses ($\mathbf{p}_i$s and $R_i$s) from $m$ relative noisy observation between poses. Since all measurements in PGO are relative, equal rotating and moving of all vertices is ineffective. Therefore, it can be assumed that v0 is the origin ($\vp_0=[0,0,0]^T$ and $R_0=I_3$). We can see poses and relative observation between poses as a directed graph $G=(V,E)$, where $V=\{v_0,v_1,...,v_n\}$ is a set of $n+1$ vertices representing poses and $E=\{e_1,...,e_m\}$ is a set of edges representing relative measurement between poses. The direction of the edge $e_k=(i,j),\quad k \in \{1,\hdots,m\}$ shows that relative observation is from node $i$ to node $j$.





The incidence matrix $\bar{A}$ of aforementioned directed graph is a $m\times(n+1)$ matrix in which $k^{th}$ row is related to edge $e_k=(i,j)$. In $k^{th}$ row of $\bar{A}$ there is a `$+1$' in $j^{th}$ column and a `$-1$' in the $i^{th}$ column and other entries of $k^{th}$ row are zero. The full column rank matrix $A$ is formed by eliminating first column of $\bar{A}$ \cite{bapat2010graphs}.

\subsection{Some Operators}
In this paper, $Sym(X)$ indicates the symmetric part, and $Skew(X)$ indicates the skew-symmetric part of the matrix $X$.
Given a 3-vector $\mathbf{a}^T=(a_x,a_y,a_z)$, the $S:\,\mathbb{R}^3\rightarrow\mathbb{R}^{3\times 3}$ operator on this vector gives a skew-symmetric matrix and is defined by
\begin{equation}\label{S}
    S(\mathbf{a}) =
    \begin{bmatrix}
      0 & -a_z & a_y \\
      a_z & 0 & -a_x \\
      -a_y & a_x & 0
    \end{bmatrix}
\end{equation}%
A 3-vector $\vm=(m_x,m_y,m_z)$ can represent the skew-symmetric part of a matrix $M$:
\begin{equation}
    Skew(M)=
    \begin{bmatrix}
      0&-m_z&m_y \\
      m_z&0&-m_y \\
      -m_y&m_x&0
    \end{bmatrix}
\end{equation}
The operator $\vs:\,\mathbb{R}^{3\times 3}\rightarrow\mathbb{R}^3$ inputs a matrix and returns the 3-vector representing skew-symmetric part of the matrix.

\subsection{Rotation Matrices and Rodrigues' Formula}
According to Euler's rotation theorem in three-dimensional space, two coordinate system with common origin can be matched by a single rotation of one of them about an axis. This axis is called Euler's axis of rotation.

Rodrigues' rotation formula gives a method for computing rotation matrix from Euler's axis and rotation angle. Assume that $\mathbf{a}$ is a unit vector in Euler's axis direction and $\theta$ is the rotation angle. According to Rodrigues' rotation formula, the rotation matrix is obtained by
\begin{equation}\label{Rodrigues}
    R = I-J\sin{\theta}+J.J(1-\cos{\theta})
\end{equation}
where $I$ is $3\times3$ identity matrix and $J=S(\mathbf{a})$.

By defining $\bm{\delta}\triangleq \mathbf{a}\sin{\theta}$, the Rodrigues' formula can be re-expressed as bellow
\begin{equation}\label{Rodrigues2}
    R = I-S(\bm{\delta})+\beta(\|\bm{\delta}\|) S^2(\bm{\delta})
\end{equation}
where
\begin{equation}\label{beta}
    \beta(\|\delta\|) = \left(\frac{1-\sqrt{1-\|\bm{\delta}\|^2}}{\|\bm{\delta}\|^2}\right)=
    \left(\frac{1-\cos{\theta}}{\sin^2{\theta}}\right)
\end{equation}
\subsection{Measurements}
Each edge in the graph indicates relative pose measurements. Relative pose measurements between vertices $i$ and $j$ can be represented as below
\begin{align}\label{measures}
    Z_i^j &= R_i^jR_{noise} = R_jR_i^TR_{noise}\\
    \mathbf{d}_{ij} &= R_i(\mathbf{p}_j-\mathbf{p}_i)+\mathbf{d}_{noise}
\end{align}

\subsection{PGO Formulation}
Assume that the directed graph $G(V,E)$ is given. As mentioned previously, PGO problem is to find the position and orientation of the vertices of the graph, i.e. $\mathbf{p}_i,R_i,i\in\{1,...,n\}$, subject to minimize the following cost function:
\begin{equation}\label{cost}
    \sum_{e_k=\{i,j\}\in E} \lambda_k\left\|\mathbf{d}_{ij}-R_i(\mathbf{p}_j-\mathbf{p}_i)\right\|^2
    +\omega_k\left\|Z_i^j-R_jR_i^T\right\|_F^2
\end{equation}
Suppose that $R_i^*,\,i=\{1,...,n\}$ are orientations of optimal solution of \eqref{cost}. Then optimal positions minimize the cost function \eqref{costp} (the first part of cost function \eqref{cost})
\begin{equation}\label{costp}
    \sum_{e_k=\{i,j\}\in E} \lambda_k\left\|\mathbf{d}_{ij}-R_i^*(\mathbf{p}_j-\mathbf{p}_i)\right\|^2
\end{equation}

On the other hand, the second part of \eqref{cost} is independent from positions, and can be minimized separately. Consequently we can minimize \eqref{costr} to find orientations and then use them to minimize \eqref{costp}.
\begin{equation}\label{costr}
    \sum_{e_k=\{i,j\}\in E} \omega_k\left\|Z_i^j-R_jR_i^T\right\|_F^2
\end{equation}
It is clear that solutions of \eqref{costr} are not exactly optimal orientations of \eqref{cost}, but they can be used as an initialization point for gradient solvers.

The problem \eqref{costp} is a linear least squares problem and has a closed form solution \eqref{solvep}
\begin{equation}\label{solvep}
    P^* = (A^T\Lambda A)^{-1}A^T\Lambda D
\end{equation}
where $P_{n\times 3}= [p_1 \dots p_n]^T$ is the matrix of positions, $A_{m\times n}$ is incidence matrix, $\Lambda_{m\times m}=diag([\lambda_1\dots\lambda_m])$ is the diagonal matrix of weights, and $D_{m\times 3}$ is the matrix of all relative displacement measurements in the reference coordinate system ($k^{th}$ row of $D$ is $d_{ij}^TR_i$ where $e_k=(i,j)\in E$). 

\section{Least Squares Approximation of the Orientation Sub-Problem}\label{Sec:LS}
Minimizing positions cost function \eqref{costp} is a linear least squares problem and can be solved easily. In this part we propose an approximation of orientation sub-problem that can be converted to a linear least squares problem.
Suppose that there are initial guesses $\hat{R}_i,\,i\in\{1,...,n\}$ for orientations. Subsequently there are rotations, $\Psi_i,\,i\in\{1,...,n\}$ that align initial guesses to vertices orientations through \eqref{deltaR}
\begin{equation}\label{deltaR}
    R_i = \hat{R}_i\Psi_i
\end{equation}

Using Rodrigues' formula, $\Psi_i$ can be written as
\begin{equation}\label{deltaR_R}
    \Psi_i = I - S(\bm{\delta}_i)+\beta_i(\|\bm{\delta}_i\|) S^2(\bm{\delta}_i)
\end{equation}

Substituting \eqref{deltaR} and \eqref{deltaR_R} in orientation sub-problem \eqref{costr}, changes the cost function arguments to $\bm{\delta}_i,\,i\in\{1,...,n\}$. Thus, orientation sub-problem can be rewritten as below.
{\fontsize{9}{1}
\begin{align}\label{costr_R}
    &\argmin_{R_1\ldots R_n}\sum_{e_k=\{i,j\}\in E}
    {\omega_{k}\left\|Z_i^j-R_jR_i^T\right\|_F^2}\nonumber\\
    &\equiv\argmin_{\Psi_1\ldots \Psi_n}{\sum_{e_k=\{i,j\}\in E}
    \omega_{k}\left\|Z_i^j-\hat{R}_j\Psi_j\Psi_i^T\hat{R}_i^T\right\|_F^2}\nonumber\\
    &\equiv\argmin_{\Psi_1\ldots \Psi_n}{\sum_{e_k=\{i,j\}\in E}
    \omega_{k}\left\|\hat{R}_j^TZ_i^j\hat{R}_i-\Psi_j\Psi_i^T\right\|_F^2}\nonumber\\
    &\equiv\argmin_{\bm{\delta}_1\ldots \bm{\delta}_n}\sum_{e_k=\{i,j\}\in E}
    \omega_{k}\Big\|\hat{R}_j^TZ_i^j\hat{R}_i-\nonumber\\
    &\quad
    \left(I-S(\bm{\delta}_j)+\beta_jS^2(\bm{\delta}_j)\right)
    \left(I+S(\bm{\delta}_i)+\beta_iS^2(\bm{\delta}_i)\right)\Big\|_F^2\nonumber\\
    &\equiv\argmin_{\bm{\delta}_1\ldots \bm{\delta}_n}\sum_{e_k=\{i,j\}\in E}
    \omega_{k}\Big\|\hat{R}_j^TZ_i^j\hat{R}_i-I+S(\bm{\delta}_j)
    -S(\bm{\delta}_i)+\nonumber\\
    &\quad
    S_2(\bm{\delta}_i,\bm{\delta}_j)\Big\|_F^2
\end{align}}
where $S_2(\bm{\delta}_i,\bm{\delta}_j)$ contains all terms with multiplication of at least two skew-symmetric terms, $S(\bm{\delta}_i)$ or $S(\bm{\delta}_j)$.
\begin{align*}
    S_2(\bm{\delta}_i,\bm{\delta}_j) &= S(\bm{\delta}_j)S(\bm{\delta}_i)\\ 
    &+S(\bm{\delta}_j)\beta_iS^2(\bm{\delta}_i)
    -\beta_jS^2(\bm{\delta}_j)S(\bm{\delta}_i)\\
    &-\beta_jS^2(\bm{\delta}_j)\beta_iS^2(\bm{\delta}_i)
    -\beta_jS^2(\bm{\delta}_j)-\beta_iS^2(\bm{\delta}_i)
\end{align*}

By removing $S_2(\bm{\delta}_i,\bm{\delta}_j)$ from \eqref{costr_R}, orientation sub-problem can be approximated by \eqref{costr_A}.
\begin{align}\label{costr_A}
    \argmin_{\bm{\delta}_1\ldots \bm{\delta}_n}\sum_{e_k=\{i,j\}\in E}
    \omega_{k}\left\|\hat{R}_j^TZ_i^j\hat{R}_i-I+S(\bm{\delta}_j)
    -S(\bm{\delta}_i)\right\|_F^2
\end{align}

By defining $B_k \triangleq I-\hat{R}_j^TZ_i^j\hat{R}_i$ and substituting in \eqref{costr_A}, approximated orientation sub-problem is rewritten as
\begin{align}\label{costr_A2}
    \argmin_{\bm{\delta}_1\ldots \bm{\delta}_n}\sum_{e_k=\{i,j\}\in E}
    \omega_{k}\left\|S(\bm{\delta}_j)
    -S(\bm{\delta}_i)-B_k\right\|_F^2
\end{align}

\begin{Proposition}\label{prop1}
    Let $A_{n\times n}$ be an arbitrary squared matrix, then
    \begin{equation*}
        \|A\|_F^2=\|skew(A)\|_F^2+\|sym(A)\|_F^2
    \end{equation*}
\end{Proposition}

According to Proposition \ref{prop1}, we have
\begin{multline}\label{costr_ls}
    \argmin_{\bm{\delta}_1\ldots \bm{\delta}_n} \sum_{e_k=\{i,j\}\in E}\omega_k\left\|S(\bm{\delta}_j)-S(\bm{\delta}_i)-B_k\right\|_F^2\\
    =\argmin_{\bm{\delta}_1\ldots \bm{\delta}_n} \sum_{e_k=\{i,j\}\in E}\omega_k\left\|Skew\left(S(\bm{\delta}_j)-S(\bm{\delta}_i)-B_k\right)\right\|_F^2\\
    +\omega_k\left\|Sym\left(S(\bm{\delta}_j)-S(\bm{\delta}_i)-B_k\right)\right\|_F^2\\
    =\argmin_{\bm{\delta}_1\ldots \bm{\delta}_n} \sum_{e_k=\{i,j\}\in E}\omega_k\left\|Skew\left(S(\bm{\delta}_j)-S(\bm{\delta}_i)-B_k\right)\right\|_F^2\\
    +\omega_k\left\|Sym\left(-B_k\right)\right\|_F^2\\
    =\argmin_{\bm{\delta}_1\ldots \bm{\delta}_n} \sum_{e_k=\{i,j\}\in E}\omega_k\left\|Skew\left(S(\bm{\delta}_j)-S(\bm{\delta}_i)-B_k\right)\right\|_F^2\\
    =\argmin_{\bm{\delta}_1\ldots \bm{\delta}_n} \sum_{e_k=\{i,j\}\in E}\omega_k\left\|S(\bm{\delta}_j)-S(\bm{\delta}_i)-\vs(B_k)\right\|_F^2\\
    =\argmin_\Delta \left\{\text{tr}\left((A\Delta-B)^T\Omega(A\Delta-B)\right)\right\}
\end{multline}
where $A_{m\times n}$ is the reduced incidence matrix and
\begin{equation}
    B =
    \begin{bmatrix}
      \vs(B_1)^T \\
      \vdots \\
      \vs(B_m)^T
    \end{bmatrix},\,
    \Delta =
    \begin{bmatrix}
      \bm{\delta}_1^T \\
      \vdots \\
      \bm{\delta}_n^T
    \end{bmatrix},\,
    \Omega = \text{diag}\left(
    \begin{bmatrix}
    \omega_1\\
    \vdots\\
    \omega_m
    \end{bmatrix}\right)\label{Omega}
\end{equation}



The problem \eqref{costr_ls} has the closed form solution
\begin{equation}\label{solver}
    \left(A^T\Omega A\right)^{-1}A^T\Omega B
\end{equation}

%


\section{Algorithm}\label{Sec:Alg}
In this section we propose algorithm \ref{algorithm1} to find vertices parameters. In the initialization step, vertices' orientations are roughly approximated using chordal method adopted by Martinec and Pajdla \cite{martinec2007robust}. Approximated orientations are used as initial guesses. The matrix $B$ is created and then least squares problem \eqref{costr_ls} is solved using equation \eqref{solver} to calculate $\hat{\Delta}$. Each row of $\hat{\Delta}$ is used to calculate the orientation of one vertex orientation. Since the problem \eqref{costr_ls} is approximation of the orientation sub-problem, so the solution is not exactly the optimal solution. To achieve better accuracy the LS approximation problem is solved again, using the last solutions as initial guesses. The iterative process continues until convergence. Finally vertices' position are calculated using \eqref{solvep}.

\renewcommand{\algorithmicrequire}{\textbf{Input:}}
\renewcommand{\algorithmicensure}{\textbf{Output:}}
\begin{algorithm}[!ht]
\newcommand{\NewComment}[1]{ {\hfill$//$ #1}} 
  \caption{Solving Approximated Orientation Sub-Problem}
    \begin{algorithmic}[1]
	    \Require{Edges of Pose Graph ($Z_i^j,\vd_{ij},\,\forall e_k=\{i,j\}\in E$) and $\Omega$ from \ref{Omega}} 
        \Ensure{Estimated Vetrices Pose ($R_i,\vp_i,\, i\in\{1,\ldots,n\}$)}

        \State $t \gets 1$
        \State $\hat{R}_i(t) \gets$  Chordal Initialization
		\Repeat
            \State $\displaystyle \vb_k\gets \vs\left(I-\hat{R}_j(t)^TZ_i^j\hat{R}_i(t)\right),\forall e_k=\{i,j\}\in E$
            \State $B \gets [\vb_1,\ldots,\vb_m]^T$
            \State $\hat{\Delta}(t)\gets(A^T\Omega A)^{-1}A^T\Omega B$
            \For{$i\in\{1,\ldots,n\}$}
                \State $\hat{\del_i}(t)^T\gets i^{th}$ row of $\hat{\Delta}(t)$
                \State $\hat{\Psi}_i(t) \gets I - S(\hat{\del}_i(t)) + \beta_i(\|\hat{\del}_i(t)\|) S^2(\hat{\del}_i(t))$
                \State $\displaystyle\hat{R}_i(t+1)\gets \hat{R}_i(t)\hat{\Psi}_i(t)$
            \EndFor
            \State $t\gets t+1$\;
		\Until{$\max_i\|\hat{\del}_i\|>tolerance$ and $t < itr_{max}$}
        \State $R_i \gets \hat{R}_i(t), \, \forall i\in\{1,\ldots,n\}$
        \State Calculate $P$ from \ref{solvep} 
        \State $\mathbf{p}_i^{T}\leftarrow i^{th}$ row of $P, \, \forall i\in\{1,\ldots,n\}$
        \State \Return {$\mathbf{p}_i,R_i, \, \forall i\in\{1,\ldots,n\}$}
    \end{algorithmic}
  \label{algorithm1}
\end{algorithm}

%

\section{Convergence Analysis}\label{Sec:CA}
In this section, we first prove the convergence of the proposed algorithm for noise-free cases. Then we find an upper bound for the distance of convergence point of the algorithm to global optimal answer.
The convergence of the method in noise-free case is not in itself worthwhile, but the proof relations can show how the algorithm moves towards the global optimum.

\subsection{Noise-free Cases}

In each step of iterative algorithm, $\hat{\Delta}$ is obtained by solving the approximated orientation sub-problem.
\begin{multline}\label{deltab}
    \hat{\Delta}(t) = \argmin\sum_{e_k=\{i,j\}\in E} \omega_k\Big\|\hat{R}_j^T(t) Z_i^j \hat{R}_i(t) - I \\
    + S(\del_j) - S(\del_i) \Big\|_F^2
\end{multline}
and then $\hat{\Psi}_i(t),\; i\in{1,...,n}$ are obtained from
\begin{align}
    \hat{\Psi}_i(t) = I - S(\hat{\del}_i(t)) + \beta_i(\|\hat{\del}_i(t)\|) S^2(\hat{\del}_i(t)) \label{DR}
\end{align}
where $\hat{\del}_i(t)$ is the $i^{th}$ row of $\hat{\Delta}(t)$. Then the estimation of orientations are updated through
\begin{equation}\label{upd}
\hat{R}_i(t+1) = \hat{R}_i(t)\hat{\Psi}_i(t)
\end{equation}

Assume that $\Delta^\star$ is the minimizer of the cost function \eqref{costr_R}.
\begin{multline}\label{deltas}
    \Delta^\star(t) = \argmin\sum_{e_k=\{i,j\}\in E} \omega_k\Big\|\hat{R}_j^T(t) Z_i^j \hat{R}_i(t) - I + \\
    S(\del_j) - S(\del_i) + S_2(\del_i,\del_j)\Big\|_F^2
\end{multline}

In noise-free cases, $Z_i^j=R_i^j$ and therefore the minimum value of cost function is zero. So $\Delta^\star(t)$ is also the solution of problem \eqref{deltas2}
\begin{multline}\label{deltas2}
    \Delta^\star(t) = \argmin\sum_{e_k=\{i,j\}\in E} \omega_k\Big\|\hat{R}_j^T(t) R_i^j \hat{R}_i(t) - I + \\
    S(\del_j) - S(\del_i) + S_2(\del_i^\star(t),\del_j^\star(t)) \Big\|_F^2
\end{multline}\

{\color{black} Note that eq. \eqref{deltas2} is not a way to calculate $\Delta^\star(t)$, but rather to express a relation for $\Delta^\star(t)$.}

The relationship between optimal orientations and $\Psi^\star(t)$ is
\begin{equation}\label{DRs}
R_i^\star = \hat{R}_i(t)\Psi^\star_i(t)
\end{equation}

Equation \eqref{Rs} is obtained by substituting \eqref{upd} into \eqref{DRs}
\begin{equation}\label{Rs}
R_i^\star = \hat{R}_i(t+1)\hat{\Psi}_i^T(t)\Psi^\star_i(t)
\end{equation}

Thus
\begin{equation}\label{Rs_upd}
\Psi^\star_i(t+1) = \hat{\Psi}_i^T(t)\Psi^\star_i(t)
\end{equation}


With respect to relations in \eqref{costr_ls}, solutions of \eqref{deltas2} and \eqref{deltab} can be obtained as follows
\begin{multline}
    \Delta^\star(t) = \argmin \Big\{\text{tr} \Big((A\Delta-(B(t)+C(t)))^T\Omega\\
    (A\Delta-(B(t)+C(t)))\Big)\Big\} \label{Apds}
\end{multline}
\begin{multline}
    \hat{\Delta}(t) = \argmin \left\{\text{tr}\Big((A\Delta-B(t))^T\Omega(A\Delta-B(t))\Big)\right\} \label{Apdb}
\end{multline}
where
\begin{align}
    B(t)^T &= \left[ \vb_1(t),\ldots,\vb_m(t) \right]\label{B}\\
    C(t)^T &= \left[ \vc_1(t),\ldots,\vc_m(t) \right]\label{C}
\end{align}
and for each $e_k=\{i,j\}\in E$
\begin{align}
    \vb_k(t) &= \vs \left(I-\hat{R}^T_j(t)R_i^j\hat{R}_i(t)\right)\\
    \vc_k(t) &= \vs \left(-S_2\left(\del_i^\star(t),\del_j^\star(t)\right)\right)
\end{align}

Problems \eqref{Apds} and \eqref{Apdb} are standard linear least squares problems. Hence their solutions are as follows
\begin{align}
    \Delta^\star(t) &= A^\dag(B(t)+C(t)) \label{Delstar}\\
    \hat{\Delta}(t) &= A^\dag(B(t))
\end{align}
where
\begin{align}\label{ddiff}
    A^\dag = \left(A^T\Omega A\right)^{-1}A^T\Omega
\end{align}

Therefore the difference between solutions is obtained as \eqref{Dif}
\begin{align}\label{Dif}
    \Delta^\star(t) - \hat{\Delta}(t) = \left(A^T\Omega A\right)^{-1}A^T\Omega C(t) = A^\dag C(t)
\end{align}
and therefore
\begin{align}\label{dif}
    \del_k^\star(t) - \hat{\del}_k(t) = \va_k^\dag(t) C(t), \quad k\in \{1,\ldots,n\}
\end{align}
where $\va_k^\dag(t)$ represents the $k^{th}$ row of $A^\dag$.


To demonstrate the convergence of the algorithm to optimal solution ($\hat{R}_i\rightarrow R^\star_i,\; i\in{1,...,n}$), it should be shown that in each step, $\Psi^\star_i$ becomes closer to identity or equally $\left\|\Delta^\star\right\|\rightarrow0$.

\begin{figure}
    \centering
    \includegraphics[width=0.4\textwidth,clip,keepaspectratio]{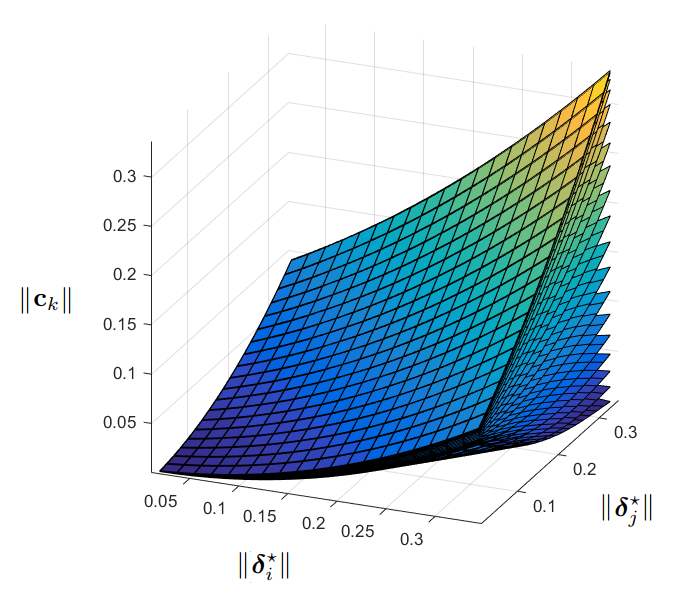}
    \caption{Each surface corresponds to a specific $\alpha$. The lowest surface corresponds to $\alpha=0^\circ$ and the highest surface corresponds to $\alpha=180^\circ$}\label{fig:db}
\end{figure}

The magnitude of $\vc_k$ is plotted according to the magnitude of $\del^\star_i$, $\del^\star_j$ and $\alpha$ (the angle between $\del^\star_i$ and $\del^\star_j$) in Fig. \ref{fig:db}. As illustrated in the figure, the maximum magnitude of $\vc_k$ occur when $\|\del^\star_i\| = \|\del^\star_j\|$ and $\del^\star_i, \del^\star_j$ are exactly at the opposite directions.

According to eq. \eqref{dif}, it can be concluded that:
\begin{equation}\label{limdd}
    \left\|\del^\star_k(t)-\hat{\del}_k(t)\right\|\leq\left\|\va_k^\dag\right\|\max_{k'}\left\|\vc_{k'}(t)\right\|
    \leq a_m c_m(t)
\end{equation}
where $c_m(t)=\max_{k}\left\|\vc_{k}(t)\right\|$ and $a_m=\max_{k}\left\|\va^\dag_{k}\right\|$.

The magnitude of $\va_k^\dag,\; k\in {1,...,n}$ depends on graph topology and is usually small (less than 2 for torus, sphere-a and cube in \cite{carlone2015initialization} datasets).
Below, the convergence area with the assumption $a_m=1$ is computed. Then the effect of $a_m$ on the obtained convergence area will be investigated.

\begin{figure}
    \centering
    \includegraphics[width=0.3\textwidth,clip,keepaspectratio]{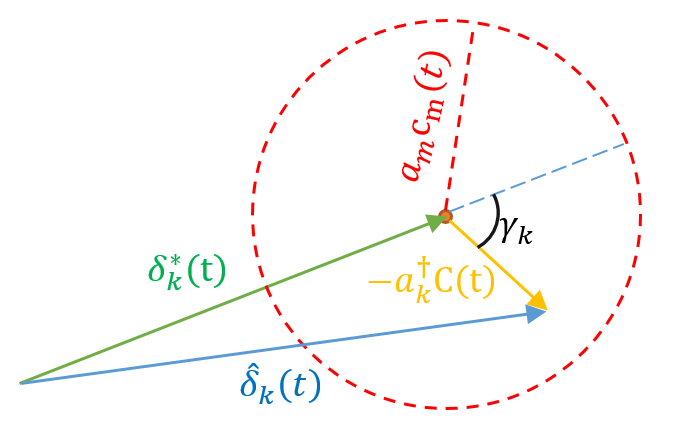}
    \caption{The magnitude of difference between $\del_k^*(t)$ and $\hat{\del}(t)$ is smaller or equal to $a_m c_m(t)$. Therefor the geometric location of $\hat{\del}_k(t)$ falls inside the red circle.}\label{fig:delta}
\end{figure}

\begin{figure}
    \centering
    \includegraphics[width=0.48\textwidth,clip,keepaspectratio]{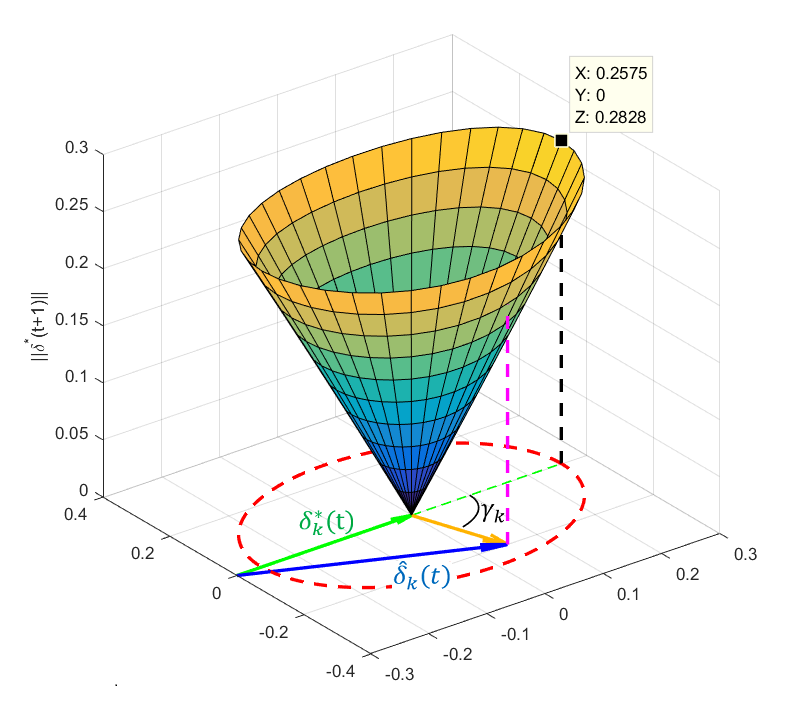}
    \caption{Using eq.\eqref{Rs_upd}, the magnitude of $\del^*_k(t+1)$ can be calculated for all points inside the geometric location of $\hat{\del}_k(t)$}\label{fig:gamma}
\end{figure}

According to \eqref{limdd}, the geometric location of $\hat{\del}_k(t)$ falls inside the red circle \footnote{Vectors $\hat{\del}_k(t)$ and $\del_k^\star(t)$ are 3-vectors and therefore, the geometric location is a sphere. But given that the rotation of $\hat{\del}_k(t)$ around $\del_k^\star(t)$ has no effect on the analysis, only the planar geometry is considered which is a circle.} in Fig. \ref{fig:delta}. The magnitude of $\del_k^\star(t+1)$ for this area is plotted using Eq.\eqref{Rs_upd} assuming $\|\del_k^\star(t)\|=0.3$ and $c_m=0.26$ (In Fig. \ref{fig:gamma}).
As illustrated in the same figure, the maximum magnitude of $\del^\star_k(t+1)$ occurs when $\gamma_k$ (the angle between $\del^\star_k(t)$ and $-\va^\dag_kC(t)$) is zero ($\del^\star_k$, $\hat{\del}_k$ and $-\va^\dag_kC(t)$ have the same direction). The general form of the shape and extremum points are the same for different values of $\|\del_k^\star(t)\|$ and $c_m$. Also bigger values for $\|\del_k^\star(t)\|$ result in bigger values for $\|\del^\star_k(t+1)\|$.
Assume that $k'=\argmax_k \|\del^\star_k(t)\|$, and $\gamma_{k'}=0$, then it can be concluded that $\|\del^\star_{k'}(t+1)\|\geq\|\del^\star_{k}(t+1)\|, k\in\{1,\ldots,n\}$. Therefore the maximum magnitude of $\del^\star_k(t+1)$ occurred for $k=k'$.

\begin{figure}
    \centering
    \includegraphics[width=0.4\textwidth,clip,keepaspectratio]{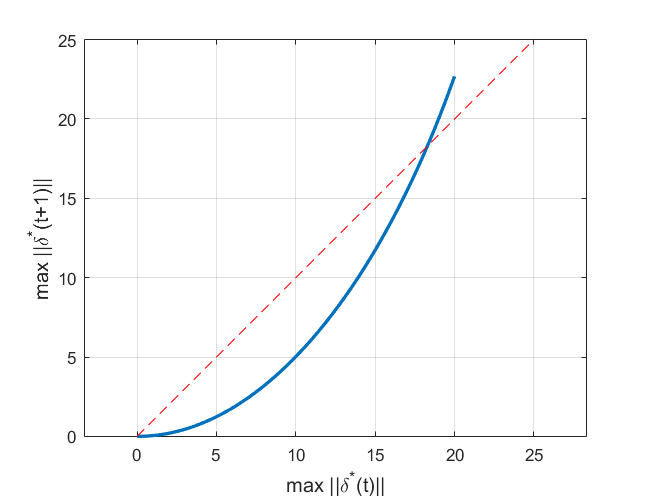}
    \caption{The magnitude of $\del^*_k(t+1)$ with respect to $\del^*_k(t)$ (Blue, Solid) and Bisector of first quadrant (Red, Dashed)}
    \label{fig:cnvg}
\end{figure}

Based on worst conditions, the maximum bound of $\del^\star_k(t+1)$ is plotted with respect to the magnitude of $\del^\star_k(t)$ in Fig.\ref{fig:cnvg}. Obviously, if $\|\del^\star_k(t)\|$ is less than a certain value, $\|\del^\star_k(t+1)\|$ will always be less, and so the algorithm converges to optimal solution.

\begin{figure}
    \centering
    \includegraphics[width=0.48\textwidth,clip,keepaspectratio]{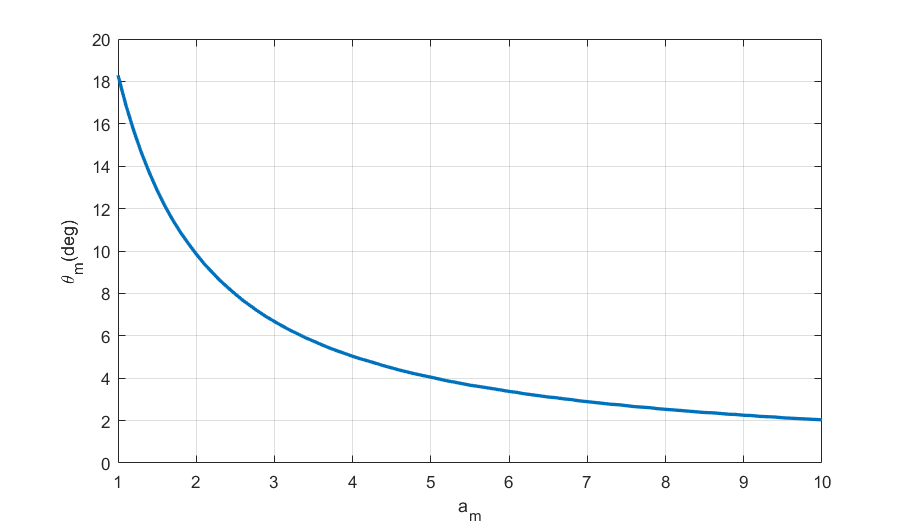}
    \caption{Maximum orientations estimation error required to prove convergence with respect to $a_m$ coefficient}
    \label{fig:am}
\end{figure}

Different values of $a_m$ change the radius of red circle in Fig. \ref{fig:delta}. A bigger $a_m$ results in larger radius for the circle and therefore bigger $\del^\star(t+1)$. Finally it results in smaller convergence area.
Fig. \ref{fig:am} shows the effect of $a_m$ on convergence area. In this figure, vertical axis is the maximum initial estimation error that ensures the convergence. For better understanding, estimation error is represented by angle of rotation (degrees).

\subsection{Noisy Cases}
In noisy cases, $\Phi_i^j$ is defined as the difference between the noisy measurement $Z_i^j$ and the real relative orientation $R_i^j$.
\begin{equation}\label{phi}
    \Phi_i^j \triangleq Z_i^j - R_i^j
\end{equation}

Suppose that $\Psi^\star_i(t)$ rotates $\hat{R}_i(t)$ to real orientation $R^\star_i$ (optimal solution in noise free assumption). Using Rodrigues formula
\begin{equation}\label{psi}
    \Psi^\star_i(t) = I - S(\del^\star_i(t))+\beta(\|\del^\star_i(t)\|) S^2(\del^\star_i(t))
\end{equation}

$\del^\star_i(t), i\in\{1,\ldots,n\}$ are the solutions of \eqref{deltas2} and can be calculated through \eqref{Delstar}.

In each step of presented algorithm $\hat{\Delta}$ is calculated through \eqref{deltabn1}
\begin{multline}\label{deltabn1}
    \hat{\Delta}(t) = \argmin\sum_{e_k=\{i,j\}\in E} \omega_k\Big\|\hat{R}_j^T(t) Z_i^j \hat{R}_i(t) - I +\\
    S(\del_j) - S(\del_i) \Big\|_F^2
\end{multline}

According to \eqref{phi}, the equation \eqref{deltabn1} can be written as follows
\begin{multline}\label{deltabn2}
    \hat{\Delta}(t) = \argmin\sum_{e_k=\{i,j\}\in E} \omega_k\Big\|\hat{R}_j^T(t) R_i^j \hat{R}_i(t) + \hat{R}_j^T(t) \Phi_i^j \hat{R}_i(t) - I +\\
    S(\del_j) - S(\del_i) \Big\|_F^2
\end{multline}

The solution of \eqref{deltabn2} is
\begin{equation}\label{}
    \hat{\Delta}(t) = A^\dag(B(t)+E(t))
\end{equation}
where $B(t)$ is introduced in eq. \eqref{B} and
\begin{align}
    E(t)^T &= \left[ \ve_1(t),\ldots,\ve_m(t) \right]
\end{align}
and
\begin{equation}\label{}
    \ve_k(t) = \vs \left(-\hat{R}^T_j(t)\Phi_i^j\hat{R}_i(t)\right)
\end{equation}

Therefore the difference between solutions is obtained as
\begin{equation}\label{Deltabar}
    \Delta^\star(t) - \hat{\Delta}(t) = A^\dag(C(t)-E(t))
\end{equation}
and therefore
\begin{align}\label{difn}
    \del^\star_k(t) - \hat{\del}_k(t) = \va_k^\dag(t) (C(t)-E(t)), \quad k\in \{1,\ldots,n\}
\end{align}
where $\va_k^\dag(t)$ represents the $k^{th}$ row of $A^\dag$.

The magnitude of the difference between solutions satisfy the following inequality.
\begin{multline}\label{ineqdif}
    \left\|\del^\star_k(t)-\hat{\del}_k(t)\right\|\leq\left\|a_k^\dag\right\|(\max_{k'}\left\|C_{k'}(t)\right\|+\max_{k''}\left\|E_{k''}(t)\right\|)\\
    \leq a_m (c_m(t)+e_m(t))
\end{multline}
where $a_m=\max_{k}\left\|\va^\dag_{k}\right\|$, $c_m(t)=\max_{k}\left\|\vc_{k}(t)\right\|$ and $e_m(t)=\max_{k}\left\|\ve_{k}(t)\right\|$.

\begin{figure}
    \centering
    \includegraphics[width=0.4\textwidth,clip,keepaspectratio]{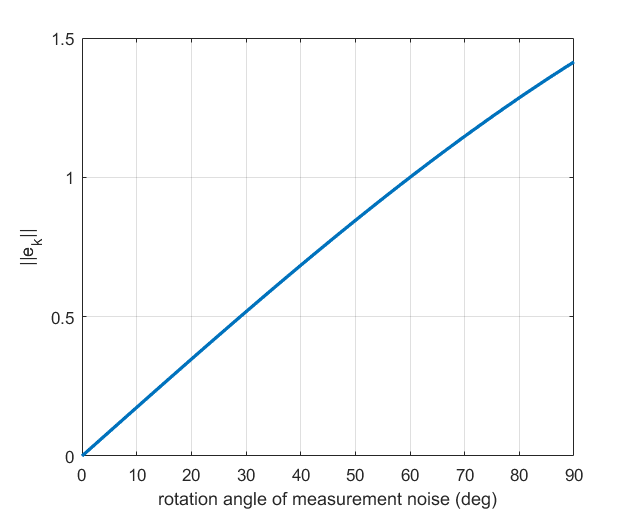}
    \caption{Assuming orientation measurement noise as a rotation described in Eq. \eqref{measures}, the magnitude of $e_k$ will only be the function of angle of this rotation}\label{fig:ek}
\end{figure}

The magnitude of $\ve_k$ only depends on the measurement noise. Fig. \ref{fig:ek} indicates the magnitude of $\ve_k$ with respect to relative orientation measurement noise. The measurement noise is represented by angle of rotation (degrees).

\begin{figure}
    \centering
    \includegraphics[width=0.4\textwidth,clip,keepaspectratio]{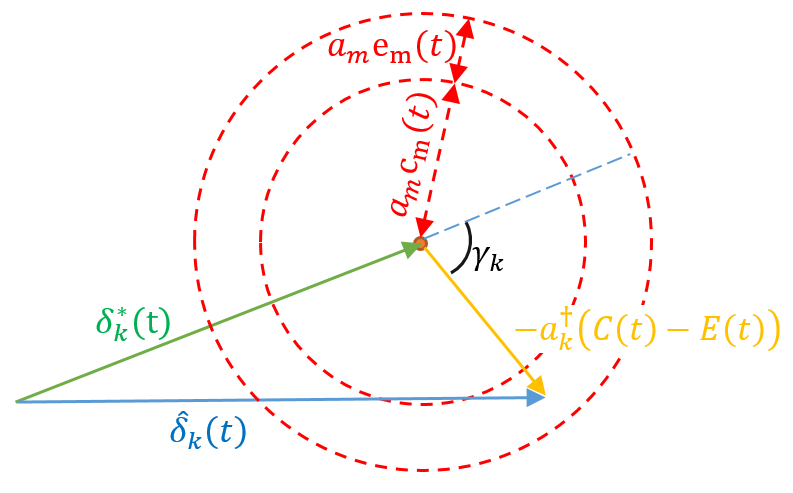}
    \caption{The radius of geometric location of $\hat{\del}_k(t)$ will be larger for noisy cases.}\label{fig:cirn}
\end{figure}

According to \eqref{ineqdif}, the geometric location of $\hat{\del}_k(t)$ falls inside the bigger red circle in Fig. \ref{fig:cirn}.

\begin{figure}
    \centering
    \includegraphics[width=0.4\textwidth,clip,keepaspectratio]{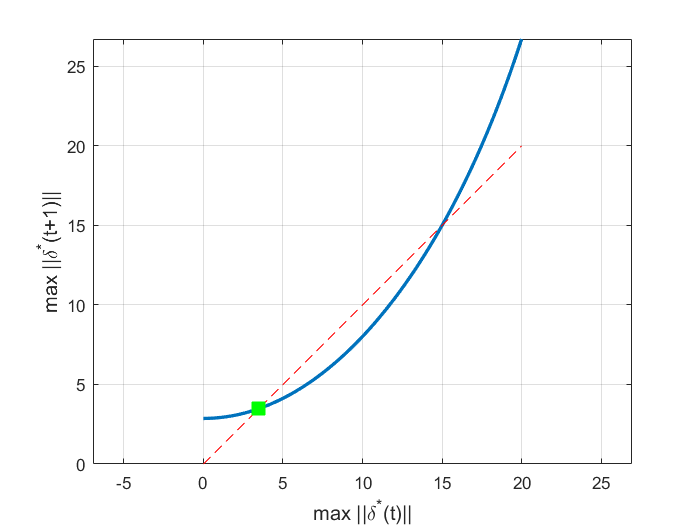}
    \caption{If the magnitude of $\del^\star(t)$ be less then $15$ degrees and be larger than $3.7$ degrees (green square), then it gets smaller in each iteration.}\label{fig:cnvgn}
\end{figure}

The magnitude of $\del_k^*(t+1)$ is the same as in Fig. \ref{fig:gamma}, with the difference that the radius of red circle is larger as much as $a_me_m$.
Like noise-free conditions, the largest value for $\|\del_k^*(t+1)\|$ is obtained when $\gamma_k$ is zero and $k = k', k'=\argmax_k \|\del^\star_k\|$. Therefore $max_k\|\del_k^*(t+1)\|$ can be calculated with respect to $max_k\|\del_k^*(t)\|$. In Fig. \ref{fig:cnvgn} $max_k\|\del_k^*(t+1)\|$ is plotted with respect to $max_k\|\del_k^*(t)\|$ with the assumptions $a_m=1$ and $e_m=0.05$.
It can be seen from Fig. \ref{fig:cnvgn} that if initial guess of the orientations is more accurate than 15 degrees and with the aforementioned assumptions, the algorithm yields an accuracy of at least 3.7 degrees (solid green square). {\color{black} Note that this does not mean that the algorithm converges to a point with a accuracy of 3.7 degrees, because the conditions we were considering to prove were not real, but this is the least accuracy that we could prove.}

\begin{figure}
    \centering
    \includegraphics[width=0.48\textwidth,clip,keepaspectratio]{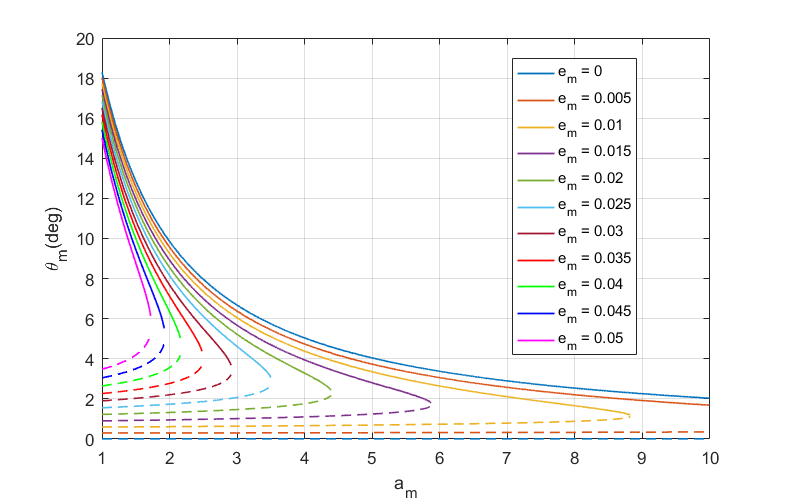}
    \caption{The effect of $e_m$ and $a_m$ on the maximum initial guess error, required to prove convergence, (solid linse) and the accuracy of algorithm (dashed lines)}\label{fig:am_em}
\end{figure}

The effect of $e_m$ and $a_m$ on the maximum initial guess error (required to prove convergence) and the accuracy of algorithm is shown in Fig. \ref{fig:am_em}

\section{Least Squares Approximation of the PGO Problem}\label{Sec:PGOLS}
The algorithm \ref{algorithm1} uses only relative orientation observations to estimate vertices orientations. In this part we propose an approximation of PGO problem which use both relative orientation and position observations and can be solved with least squares solvers. So better initialization point can be obtained which is very close to the optimal solution in low noise cases. We use Rodrigues’ rotation formula by removing the second order term and substituting in original PGO problem \eqref{cost}.
\begin{multline}\label{cost_A}
    \min_{\del_i,\mathbf{p}_i}\sum_{e_k\in E}
    \lambda_{ij}\left\|\mathbf{p}_j-\mathbf{p}_i-\hat{R}_i^T\mathbf{d}_{ij} - S(\bm{\delta_i})\hat{R}_i^T\mathbf{d}_{ij}\right\|^2+\\
    \omega_{ij}\left\|\hat{R}_j^TZ_i^j\hat{R}_i-I+S(\del_j)
    -S(\del_i)\right\|_F^2\\
    \equiv
    \min_{\del_i,\mathbf{p}_i}\sum_{e_k\in E}
    \lambda_{ij}\left\|\mathbf{p}_j-\mathbf{p}_i-\hat{R}_i^T\mathbf{d}_{ij} + S\left(\hat{R}_i^T\mathbf{d}_{ij}\right)\bm{\delta_i}\right\|^2+\\
    2\omega_{ij}\left\|\del_j-\del_i-\vs\left(I-\hat{R}_j^TZ_i^j\hat{R}_i\right)\right\|^2
\end{multline}

Optimization problem can be rewritten in the matrix form
\begin{equation}\label{cost_AM}
    \min_{\del_i,\mathbf{p}_i} \left\|
    \begin{bmatrix}
      (2\Omega A)\otimes I & O \\
      M & (\Lambda A)\otimes I
    \end{bmatrix}
    \begin{bmatrix}
        \del_T\\
        \mathbf{p}_T
    \end{bmatrix}
    -
    \begin{bmatrix}
        2Z_\Omega\\
        D_\lambda
    \end{bmatrix}
    \right\|^2
\end{equation}
where
\begin{equation}
    \del_T =
    \begin{bmatrix}
        \del_1\\
        \vdots\\
        \del_n\\
    \end{bmatrix}, \,
    \mathbf{p}_T =
    \begin{bmatrix}
        \mathbf{p}_1\\
        \vdots\\
        \mathbf{p}_n
    \end{bmatrix}, \,
    Z_\Omega =
    \begin{bmatrix}
        \vz_1\\
        \vdots\\
        \vz_m
    \end{bmatrix}, \,
    D_\lambda =
    \begin{bmatrix}
        \vl_1\\
        \vdots\\
        \vl_m
    \end{bmatrix}
\end{equation}
and for each $e_k = \{i, j\} \in E$
\begin{equation}
    \vz_k = \omega_{ij}\vs\left(I-\hat{R}_j^TZ_i^j\hat{R}_i\right)
    , \quad
    \vl_k = \lambda_{ij}\hat{R}_i^T\mathbf{d}_{ij}
\end{equation}

and $\otimes$ represents Kronecker product. $M$ is a matrix with $m$ rows and $n$ columns of $3\times3$ blocks in which for each $e_k = \{i, j\} \in E$ the $i^{th}$ column-block in $k^{th}$ row-block is equal to $\lambda_{ij}S\left(\hat{R}_i^T\mathbf{d}_{ij}\right)$ and other terms are zero.

The Solution of \eqref{cost_AM} is obtained as
\begin{equation}\label{solv_AM}
    \begin{bmatrix}
        \del^*_T\\
        \mathbf{p}^*_T
    \end{bmatrix}
    =
    G^\dag
    \begin{bmatrix}
        2Z_\Omega\\
        D_\lambda
    \end{bmatrix}
\end{equation}
The matrix $G$ is a sparse matrix, therefore sparse solvers can be used to solve (\ref{cost_AM}). The equation \eqref{solv_AM} requires more calculations than \eqref{solver} but can lead to more accurate answer.

The algorithm \ref{algorithm2} uses just expressed formulation iteratively to solve PGO problem.

\renewcommand{\algorithmicrequire}{\textbf{Input:}}
\renewcommand{\algorithmicensure}{\textbf{Output:}}
\begin{algorithm}[!ht]
\newcommand{\NewComment}[1]{ {\hfill$//$ #1}} 
  \caption{Solving Approximated PGO Problem}
    \begin{algorithmic}[1]
	    \Require{Edges of Pose Graph ($Z_i^j,\vd_{ij},\,\forall e_k=\{i,j\}\in E$) and weights $\Omega=diag([\omega_1\dots\omega_m])$ and $\Lambda_{m\times m}=diag([\lambda_1\dots\lambda_m])$}
        \Ensure{Estimated Vetrices Pose ($R_i,\vp_i,\, i\in\{1,\ldots,n\}$)}

        \State $t \gets 1$
        \State $\hat{R}_i(t) \gets$  Chordal Initialization
		\Repeat
            \For{$e_k=\{i,j\}\in E$}
                \State $M_k \gets [0_{3\times3},\ldots,0_{3\times3},\overbrace{S\left(\hat{R}_i(t)^T\mathbf{d}_{ij}\right)}^{i^{th} column-block},$ \\ \hspace*{3cm} $0_{3\times3},\ldots,0_{3\times3}]$
                \State $\vz_k \gets \omega_{ij}\vs\left(I-\hat{R}_j^TZ_i^j\hat{R}_i\right)$
                \State $\vl_k \gets \lambda_{ij}\hat{R}_i^T\mathbf{d}_{ij}$
            \EndFor
            \State $M(t) \gets [M_1^T,\ldots,M_m^T]^T$
            \State $Z_\Omega \gets [\vz_1^T,\ldots,\vz_m^T]^T$
            \State $D_\lambda \gets [\vl_1^T,\ldots,\vl_m^T]^T$
            \State $G(t) \gets
            \begin{bmatrix}
              (2\Omega A)\otimes I & O \\
              M & (\Lambda A)\otimes I
            \end{bmatrix}$
            \State Solve for $\del_T(t)$ in
            $G
            \begin{bmatrix}
                \del_T(t)\\
                \mathbf{p}_T(t)
            \end{bmatrix}
            =
            \begin{bmatrix}
                2Z_\Omega\\
                D_\lambda
            \end{bmatrix}$
            \For{$i\in\{1,\ldots,n\}$}
                \State $\hat{\del_i}(t)\gets (3i)^{th}$ to $(3i+3)^{th}$ row of $\del_T(t)$
                \State $\hat{\Psi}_i(t) \gets I - S(\hat{\del}_i(t)) + \beta_i(\|\hat{\del}_i(t)\|) S^2(\hat{\del}_i(t))$
                \State $\displaystyle\hat{R}_i(t+1)\gets \hat{R}_i(t)\hat{\Psi}_i(t)$
            \EndFor
            \State $t\gets t+1$
		\Until{$\max_i\|\hat{\del}_i\|>tolerance$ and $t < itr_{max}$}
        \State $R_i \gets \hat{R}_i(t), \, \forall i\in\{1,\ldots,n\}$
        \State Calculate $P$ from \ref{solvep} 
        \State $\mathbf{p}_i^{T}\leftarrow i^{th}$ row of $P, \, \forall i\in\{1,\ldots,n\}$
        \State \Return {$\mathbf{p}_i,R_i, \, \forall i\in\{1,\ldots,n\}$}
    \end{algorithmic}
    \label{algorithm2}
\end{algorithm}

\section{Evaluation}\label{Sec:Eval}
In this section the performance of proposed algorithms is evaluated. First, the performance of the proposed algorithms are evaluated on common benchmark datasets \cite{carlone2015initialization} that we call low noise scenarios. Second, we add noise to these datasets and use the output of our algorithms as initialization point of well-known Levenberg-Marquardt method and the state-of-the-art SE-Sync method and compare them with the state-of-the-art chordal initialization algorithm to show the effectiveness of presented algorithms in high noise cases. In all experiments $itr_{max}$ and $tolerance$ are $10$ and $1e-4$ respectively for both algorithms \ref{algorithm1} and \ref{algorithm2}. All methods execute by Matlab 9.1 on a computer with an Intel Core i7-7700 at 2.8GHz, with 12GB of memory.
In this section the phrase $method1 + method2$ means using the $method1$ as initializer and the $method2$ as the solver.

\subsection{Low noise scenarios}
Table. \ref{tab:lownoise} shows the result of Algorithm\ref{algorithm1} and Algorithm\ref{algorithm2} and the state-of-the-art method SE-Sync \cite{rosen2017se} on the common datasets \cite{carlone2015initialization}. In this table, the cost function ($f^*$), the number of iterations ($itr$) and the execution time of algorithms ($time$) are presented. In all scenarios the cost obtained by SE-Sync is less than the cost obtained by Algorithm\ref{algorithm2} and Algorithm\ref{algorithm1}. But as it is clear from the table, the differences are low. On the other hand the Algorithm\ref{algorithm1} is almost ten times faster than the SE-Sync.
In applications with low observational noise level where speed is of great importance, the differentiation between the result of Algorithm\ref{algorithm1} and the optimal solution of SE-Sync may be negligible. So this algorithm can be used to obtain a good solution very fast. The Algorithm\ref{algorithm2} results in better solution than Algorithm\ref{algorithm1} but it is slower. However Algorithm\ref{algorithm2} was faster than SE-Sync in almost all scenarios.

\begin{table*}[!thbp]
    \centering
    \caption{Cost function, number of iteration and CPU time for different scenarios comparing Algorithm\ref{algorithm1}, Algorithm\ref{algorithm2} and SE-Sync method}
    \begin{tabular}{|c|c|c|c|c|c|c|c|c|c|}
        \hline
        &\multicolumn{3}{|c|}{Algorithm1}&\multicolumn{3}{|c|}{Algorithm2}&\multicolumn{3}{|c|}{SE-Sync}\\
        \hline
        Dataset   &  $f^*$  & $itr$ & $time$ &  $f^*$  & $itr$ & $time$ &  $f^*$  & $itr (outer(inner))$ & $time$ \\
        \hline
        \hline
        sphere\_a & 2963988 &   6   &  0.35  & 2963992 &   6   &   1.1  & 2961756 & 10(29)&  3.2   \\
        \hline
        torus     & 24576   &   4   &  0.63  & 24389   &  10   &   1.9  & 24252   & 4(35) &  5.15   \\
        \hline
        cube      & 86778   &   3   &  2.98  & 85333   &  10   &  20.1  & 84319   & 4(36) &  15.43  \\
        \hline
        garage    & 1.415   &   1   &  0.21  & 1.276   &  10   &   0.6  & 1.263   & 4(383)&  4.03  \\
        \hline
        cubicle   & 843.9   &   3   &  1.79  & 814.6   &  10   &   3.4  & 717.1   & 4(82) &  5.78  \\
        \hline
        rim       & 8883    &   10  &  7.19  & 7749    &  10   &   9.1  & 5460    & 5(265)&  26.16  \\
        \hline
    \end{tabular}
    \label{tab:lownoise}
\end{table*}

\subsection{High noise scenarios}
\begin{figure}
    \centering
    \includegraphics[width=0.48\textwidth,clip,keepaspectratio]{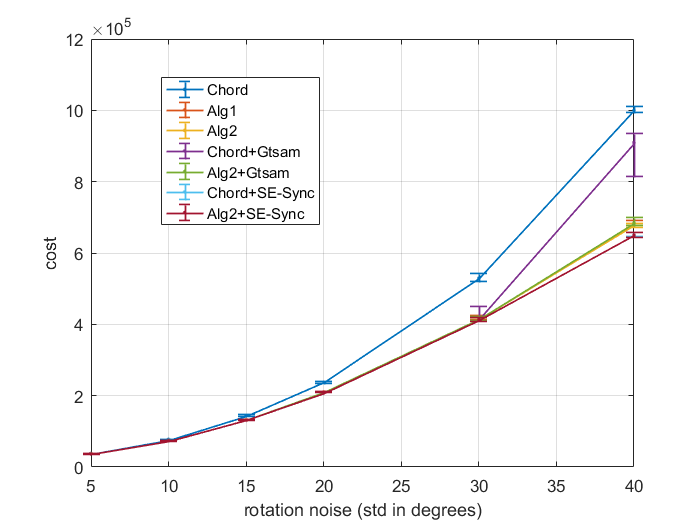}
    \caption{Cost (median and positive and negative std) VS rotation noise (std in degrees) for the torus scenario for different techniques. Each plot shows the median of ten random added noise scenario and error bars show negative and positive standard deviation.}
    \label{fig:torus}
\end{figure}

\begin{figure}
    \centering
    \includegraphics[width=0.48\textwidth,clip,keepaspectratio]{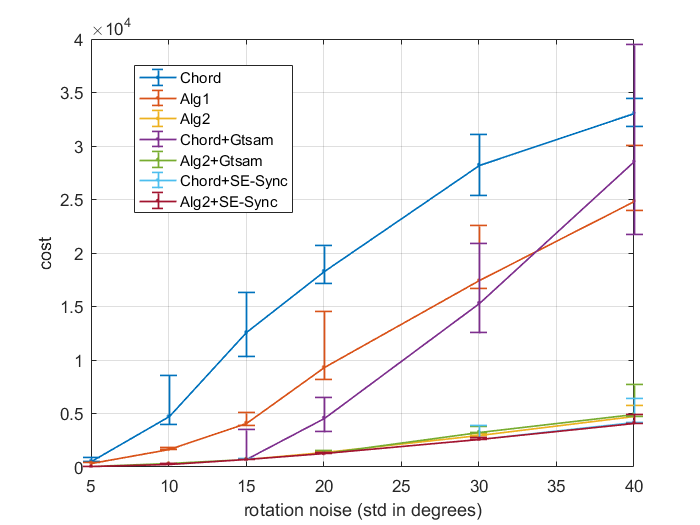}
    \caption{Cost (median and positive and negative std) VS rotation noise (std in degrees) for the garage scenario for different techniques. Each plot shows the median of ten random added noise scenario and error bars show negative and positive standard deviation.}
    \label{fig:garage}
\end{figure}

In this part the main feature of proposed algorithms is presented by adding noise to benchmark scenarios. Our algorithms are used as the initializer of Levenberg-Marquardt method (present in gtsam \cite{dellaert2012factor}) and a very stable and reliable SE-SYNC method, and the results are compared to mentioned solvers starting from chordal initialization estimation. Fig. \ref{fig:torus} and Fig. \ref{fig:garage} give an overview of costs attained by different algorithms. These figures show the cost (Eq. \ref{cost}) of chord, Alg.\ref{algorithm1}, Alg.\ref{algorithm2}, chord+gtsam, Alg.\ref{algorithm2}+gtsam, chord+SE-Sync, Alg.\ref{algorithm2}+SE-Sync methods, versus rotation noise for torus and garage datasets. As seen in Fig. \ref{fig:torus}, all methods except the chord method have almost the same results for the torus scenario with zero to thirty degrees added noise. It is clear that the solutions attained by our algorithms are much closer to optimal answer compared to chord method. This causes, in scenario with $40$ degrees noise, where the chord+gtsm entangled in local minima, algorithm Alg.\ref{algorithm2}+gtsam reaches much better solution. As shown in Fig. \ref{fig:garage}, differences between the result of various methods are obvious for garage scenario. As expected, the best result for any amount of added noise is obtained by the Alg.\ref{algorithm2}+SE-Sync. The results of Alg.\ref{algorithm2} are very close to the best result for all experiments. An important point that identifies the advantage of using our algorithms as the initializer is the difference between the results of chord+gtsam and chord+SE-Sync compared to Alg.\ref{algorithm2}+gtsam and Alg.\ref{algorithm2}+SE-Sync. The results show that the gtsam algorithm initiated by chord method, will probably stop at local minima in the case of above $20$ degrees noise in garage scenario. The amount of noise to fail to converge to a good solution for SE-Sync is about $50$ degrees.

To demonstrate the effectiveness of our initialization algorithms, we add noise to all scenarios of \cite{carlone2015initialization}. Then we compare the results of chord+SE-Sync, Alg.\ref{algorithm1}+SE-Sync and Alg.\ref{algorithm2}+SE-Sync. The results present in Table. \ref{tab:highnoise_Se}. In the table, the cost function of initialization method ($f_{init}$), the cost function of SE-Sync algorithm ($f^*$), the number of inner iterations of SE-Sync ($itr$), the execution time of SE-Sync ($time$) and the execution time of initialization algorithm ($time_{init}$) are presented. Since Alg.\ref{algorithm1} and Alg.\ref{algorithm2} also use chord, so the chord implementation time is not important in comparing the methods. The first significant point is that the cost of Alg.\ref{algorithm1} and Alg.\ref{algorithm2} is much less than the chord method in almost all scenarios. Therefore, our algorithms can provide a starting point closer to the optimal answer. This can lead to avoid local minima in challenging scenarios. Getting stuck in local minima can be seen in the table in torus and rim scenarios and is very clear about garage scenario. The slight difference between the cost of  SE-Sync, starting from Alg.\ref{algorithm1} and Alg.\ref{algorithm2} in just mentioned three scenarios, is probably due to stop conditions of SE-Sync algorithm. The second remarkable point is the effect of our algorithms on the number of SE-Sync iterations. Both Alg.\ref{algorithm1} and Alg.\ref{algorithm2}, on average on different scenarios, reduce the number of iterations about $40\%$.

In high noise scenarios execution time of Alg.\ref{algorithm1} and Alg.\ref{algorithm2} is about $40$ and $15$ times faster than SE-Sync, started from chord method.

\begin{table*}[!thbp]
    \centering
    \caption{The effect of different initialization method on the SE-Sync cost, execution time, and number of iterations}
    \begin{tabular}{|c|c|c|c|c|c|c|c|c|c|}
        \hline
        Dataset   & noise($\sigma$) & initialization method & $f_{init}$ & $f^*$ & $itr$ & $time_{init}$ & $time$ \\
        \hline
        \hline
        \multirow{3}{*}{sphere\_a} & \multirow{3}{*}{$50\degree$}
        &       chord           & $19138772$ &  $16334900$  & $34$  &    -       &  $4.56$ s \\
        &&      Alg 1           & $16347981$ &  $16334900$  &  $7$  &  $0.54$ s  &  $2.72$ s \\
        &&      Alg 2           & $16347834$ &  $16334900$  &  $7$  &  $1.45$ s  &  $3.10$ s \\
        \hline
        \multirow{3}{*}{torus} & \multirow{3}{*}{$50\degree$}
        &       chord           & $1713191$  &   $910696$   & $128$ &    -       &  $53.55$ s \\
        &&      Alg 1           & $1034042$  &   $900883$   & $134$ &  $0.70$ s  &  $61.20$ s \\
        &&      Alg 2           & $1024526$  &   $901675$   & $121$ &  $1.93$ s  &  $49.95$ s  \\
        \hline
        \multirow{3}{*}{cube} & \multirow{3}{*}{$50\degree$}
        &       chord           & $4787273$  &   $3269927$  & $16$  &    -       &  $19.72$ s \\
        &&      Alg 1           & $3378912$  &   $3269927$  &  $9$  &  $3.56$ s  &  $16.14$ s \\
        &&      Alg 2           & $3301695$  &   $3269927$  &  $8$  & $20.22$ s  &  $15.52$ s  \\
        \hline
        \multirow{3}{*}{garage} & \multirow{3}{*}{$50\degree$}
        &       chord           &  $28910$   &   $10723$    & $147$ &    -       &  $40.76$ s \\
        &&      Alg 1           &  $24974$   &    $6106$    &  $70$ &  $0.32$ s  &  $13.96$ s \\
        &&      Alg 2           &   $6888$   &    $6156$    &  $90$ &  $0.77$ s  &  $33.44$ s  \\
        \hline
        \multirow{3}{*}{cubicle} & \multirow{3}{*}{$50\degree$}
        &       chord           &  $658545$   &  $337175$   & $13$  &    -       &  $9.52$ s \\
        &&      Alg 1           &  $390476$   &  $337175$   & $13$  &  $1.91$ s  &  $9.52$ s \\
        &&      Alg 2           &  $380592$   &  $337175$   &  $7$  &  $3.69$ s  &  $6.60$ s  \\
        \hline
        \multirow{3}{*}{rim} & \multirow{3}{*}{$50\degree$}
        &       chord           &  $2152404$  &  $648068$   & $48$  &    -       &  $71.74$ s \\
        &&      Alg 1           &  $1035017$  &  $642106$   & $22$  &  $6.84$ s  &  $44.15$ s \\
        &&      Alg 2           &   $833583$  &  $641458$   & $32$  &  $9.06$ s  &  $61.34$ s  \\
        \hline
    \end{tabular}
    \label{tab:highnoise_Se}
\end{table*}

%


\section{Discussion}\label{Sec:Dis}
In this paper, we use Rodrigues rotation formula and convert the orientation sub-problem (\S\ref{Sec:LS}) and PGO problem (\S\ref{Sec:PGOLS}) into a standard least squares problem. We present two algorithms that iteratively solve LS problems to provide answers near to optimal solution. The Alg.\ref{algorithm1} is very fast and gets close to the optimal solution in datasets with reasonable amount of noise. The Alg.\ref{algorithm2} is more accurate than Alg.\ref{algorithm1}, but has more computational load.
The main feature of our algorithms is approaching to the global optimum. This can lead to escape from the local minima and offer a suitable starting point for gradient techniques. We showed that using our algorithms rather than the well-known chordal method, can be effective on improving the results, reducing the execution time and number of iterations of solvers.

In the convergence analysis we discussed the effect of coefficient $a_m$ on the convergence area and showed that bigger $a_m$ makes the convergence area smaller. This coefficient for our simulation scenarios are illustrated in Table \ref{tab:am}.
\begin{table}[!thbp]
    \centering
    \caption{Coefficient $a_m$ for different scenarios}
    \begin{tabular}{|c|c|}
        \hline
        Dataset   &  $a_m$ \\
        \hline
        \hline
        sphere\_a & 1.22 \\
        \hline
        torus     & 1.93 \\
        \hline
        cube      & 1.22 \\
        \hline
        garage    & 7.34 \\
        \hline
        cubicle   & 1.79 \\
        \hline
        rim       & 2.35 \\
        \hline
    \end{tabular}
    \label{tab:am}
\end{table}

This is important to note that the garage scenario which simulation results showed that it is more difficult to solve for all methods, has much bigger $a_m$.
Since this coefficient depends on the structure of the graph, it may be possible (we don't know how) to prune the graph in order to reduce the coefficient and make the problem easier for the solvers. Perhaps it can be used during data mining to make observations so that the resulting graph has a smaller coefficient.

A convergence area was calculated for Alg.\ref{algorithm1} in \S\ref{Sec:CA}. But in practice it is seen that the convergence area is much larger than the calculated area. Perhaps a better analysis can calculate the convergence area more accurately.



%
%
%


\bibliography{MyBiblio}{}
\bibliographystyle{IEEEtran}


\end{document}